\documentclass[letterpaper]{article} 
\usepackage{aaai25}  
\usepackage{times}  
\usepackage{helvet}  
\usepackage{courier}  
\usepackage[hyphens]{url}  
\usepackage{graphicx} 
\usepackage{svg} 
\usepackage{natbib}  
\usepackage{caption} 
\usepackage{algorithm}
\usepackage{algorithmic}

\usepackage{newfloat}
\usepackage{listings}
\DeclareCaptionStyle{ruled}{labelfont=normalfont,labelsep=colon,strut=off} 
\floatstyle{ruled}
\newfloat{listing}{tb}{lst}{}
\floatname{listing}{Listing}
\title{Towards Efficient Neurally-Guided Program Induction for ARC-AGI}

\author{
    Simon Ouellette
}

\begin{document}

\maketitle

\begin{abstract}

ARC-AGI is an open-world problem domain in which the ability to generalize out-of-distribution is a crucial quality. Under the program induction paradigm, we present a series of experiments that reveal the efficiency and generalization characteristics of various neurally-guided program induction approaches. The three paradigms we consider are \textit{Learning the grid space}, \textit{Learning the program space}, and \textit{Learning the transformation space}. We implement and experiment thoroughly on the first two, and retain the second one for ARC-AGI submission. After identifying the strengths and weaknesses of both of these approaches, we suggest the third as a potential solution, and run preliminary experiments. 

\end{abstract}

\section{Introduction}

Deep Learning is notoriously powerful when addressing a wide variety of problems. Yet, in other areas, it is deficient. One predominant difference between the problem domains where it is highly successful, and the domains where it is less so, is the openness of that domain. In closed-world domains, where the model is able to densely sample and interpolate over the full set of possbilities, deep learning often achieves near-human or sometimes even super-human ability. However, in open-world domains, where the full range of possibilities is so vast that it is practically infeasible to densely sample its entirety, interpolation-based techniques like Deep Learning can be sub-optimal.

This results in Deep Learning's well-known challenges in generalizing outside the training set distribution. The Abstraction \& Reasoning Corpus \cite{arc-prize-2024} (ARC-AGI) is one of the foremost problem domains that specifically challenges this closed-world assumption. This is done by using a hidden test set that contains tasks that are qualitatively distinct from any of the publicly available tasks. This encourages research on extending the out-of-distribution generalization capablities of learning systems. This will be the focus of the work presented here.

The current implementations of the proposed approaches are not yet mature enough to yield an interesting or competitive performance on ARC-AGI. This paper reports proof-of-concept results of experiments made under controlled and limited conditions, exploring ways in which we can efficiently extend the generalization capabilities of a Deep Learning model via search, in a program induction context.

We discuss three paradigms: \textit{Learning the grid space} (LGS), \textit{Learning the program space} (LPS), and \textit{Learning the transformation space} (LTS). The first two have been implemented and thoroughly experimented upon. The second one (LPS) is the best performing method on the ARC evaluation set, so it is retained for ARC-AGI submission. It is the approach that is therefore described in most detail. The third one (LTS) has not been implemented yet, due to lack of time, but it is described in a brief, general way. This third approach is proposed as a way to combine the strengths and weaknesses of the other two approaches. While preliminary experiments are made to support this hypothesis, full implementation and experimentation is left to future work.

\paragraph{Our main contributions} We present a few novel results:
\begin{enumerate}
    \item To the best of our knowledge, we are the first to analyze the out-of-distribution generalization characteristics of program induction approaches based on enumerating the program probability space.
    \item We present the first results on ARC-AGI for the LGS approach, and analyze its strengths and weaknesses.
    \item We propose a novel probabilistic program enumeration-based search algorithm for program induction, leaning heavily on Transformer-based auto-regressive token sequences, rather than the typical n-gram approach, and analyze its strengths and weaknesses.
    \item We sketch the outline of a novel approach (LTS) aimed at addressing the flaws in the previous approaches. We provide preliminary experimentation to support the hypotheses being advanced.
\end{enumerate}

\section{The Problem}

The problem setting we use for this paper follows a typical program synthesis setting. We assume that the Domain Specific Language (DSL) includes all the necessary primitives to solve each test task. The goal is to search (as efficiently as possible) for the program that solves each task in the test set, given the provided DSL. Throughout this paper, the tasks experimented upon are mainly taken from the ARC-AGI evaluation set.

A compute and memory time budget is allocated for each task solution, so it must find it within those budget constraints. This can be easy, even trivial for very small programs, but due to combinatorial explosion can become quite unmanageable for large programs. More formally, given a DSL $\Omega = \{\pi_1, \pi_2, ..., \pi_N\}$ containing $N$ primitive functions denoted $\pi_i$, a search algorithm $F(X, Y)$ given the support input examples $X_s$ and support output examples $Y_s$ must return within the allocated CPU and memory budgets a program $P$ such that $P(X_q) = Y_q$, where $X_q$ and $Y_q$ are the query examples. In other words, the support examples are used to infer the underlying program, and the query examples are used to confirm the correctness of the program.

There is an additional, crucial aspect to our problem setting. While we assume the DSL $\Omega$ to fully contain the required primitives to solve both the training and test datasets, we allow the test dataset's solution programs to follow different structures than the test datasets. In other words, we place the additional constraint to our problem domain that the test set must be out-of-distribution with respect to the training set in terms of the compositional structures of $P$. This is important because, if it were not the case, a neural network would be sufficient to solve this problem domain.

\section{Learning the Grid Space}

The idea of this approach is to learn a model of the space of possible ARC-AGI grids, under a specific DSL. It is important to note that there is no such thing as a pure grid similarity space unconditioned on some DSL, since under two different DSLs the same grid pair can have quite different distances. From there, it is possible to input two different grids and estimate their similarity. This can be used in a search algorithm to gradually build a program in an execution-guided manner. The idea of execution-guided program induction is to leverage some heuristic to guide the choice of primitives during construction. This heuristic requires feedback from partial execution of the program being constructed. As such, program solution construction and execution happen in parallel and impact each other. 

At each iteration of the search, the algorithm attempts all possible primitives on the current intermediate grid in the program execution, and selects the one that brings it the closest to the target grid state. Note that this approach assumes that all (or most) steps in program execution result in a grid-to-grid transformation. This is why the experiments on LGS use a DSL that strictly contains grid-to-grid transformations (DSL Version 1). As such, it should be kept in mind that the results reported for this method are therefore the best case scenario where all steps in program execution result in a grid transformation.

The experiments reported here used a Transformer encoder-only model with max pooling that outputs a flattened vector: a grid embedding. The training procedure consists of iteratively feeding a pair of grids into the model, retrieving their respective embeddings, and then penalizing deviations between their dot product and their ground truth distance.

The training data is generated by randomly sampling patches of ARC-AGI training set grids, and then picking a random number of grid transformation primitives to apply to them(zero to eight, in our expeirments). This randomly selected sequence of transformations is executed, and the output grid is retained. The input and output grids, along with the number of transformations applied, constitute a full training sample. The number of transformations is converted to a similarity between zero and one.

It should be noted that the procedurally generated training data often contained over-estimated transformation distance ground truths. This is because it is very easy to accidentally generate redundant transformations. The desirable quality of a good grid distance ground truth is that it represents the shortest distance between two grids, rather than whatever distance was arbitrarily used to generate these two grids. A trivial example of this is that "rotate 90, rotate 90, rotate 90" should yield a grid distance of one (rotate 270), not three steps. Several heuristics were implemented in the training data generation script to attempt to mitigate this issue, but nonetheless the issue remains. This is arguably one of the weaknesses of applying supervised learning to a cost-to-go type of approach.

\section{Learning the Program Space}

This is the selected approach, because it performed better than LGS on the ARC-AGI evaluation set. As such, it will be described in more depth than the other two methods. Furthermore, because we are referring to a specific, novel implementation of the overarching paradigm of LPS, we name it \textit{GridCoder}.

At a high-level, the solution consists of training a transformer to output a program, using a pre-determined grammar (DSL) and syntax, that solves the task. Specifically, the transformer outputs a sequence of probabilities over tokens that can be interpreted as a probability distribution over program space. We then use a search algorithm to enumerate that space and test valid programs for correctness. In a sense, the search explores the area of the solution space covering the neural network's region of uncertainty.

\subsection{The Search Algorithm}

It is a probability-based enumeration of programs, conceptually similar to DreamCoder \cite{dreamcoder} and Heap Search \cite{heapsearch}. However, unlike these approaches, the probability predictions are not based on n-grams, but instead are based on transformer generated token probability sequences. That is, instead of predicting the probability over children classes conditional on (n-1) parents, we predict the probability over children classes conditional on the entire maximum probability token sequence (as in typical auto-regressive transformer decoding). 

Also, unlike the other two paradigms presented in this paper (LGS and LTS), this approach is not execution-guided, in that it does not receive feedback about the intermediate state of the program as it develops a solution. Initially, there was hope that the transformer decoder could learn to do this implicitly, however the empirical results point to this not being the case (at least, not with the current architecture).

A first full decoding loop is executed on the Vision-Language Model (VLM) that is fed one input and output grid pair from a given task's demonstration set. The full decoding loop stops until either the maximum sequence length (40, in our experiments) is reached, or a token is found where the only class that has a probability greater than some threshold $\tau$ is the \texttt{End Of Sentence} class. See algorithm \ref{alg:getProbSpace} for the pseudo-code of a full decoding loop.

From there, there is a short \textit{bootstrapping} phase in which a few more inference runs are made. In these, the example pair index and the starting token (or, sometimes, the first two starting tokens) are randomly selected. The full probability distrbution is calculated as a mean of the probabilities over tokens for all of these initial decoding loops. Hence, the probability distribution $prob\_dist$ (see algorithm \ref{alg:gridcoder}) is a two-dimensional array, where one dimension represents the token positions in the sequence, and the other dimension represents the probability distribution over the possible classes (i.e. primitives in the DSL) at that token position. Therefore, our probability distribution is represented as a flat sequence, rather than a tree in which the probability of children depends on which parents are selected. This is a simplifying assumption of conditional independence, similar to the one made in Bayesian networks, that increases the efficiency of the search in our experiments.

\begin{algorithm}
\caption{GridCoder}
\label{alg:gridcoder}
\textbf{Input}: $X$, input for each example \\
\textbf{Input}: $Y$, target for each example \\
\textbf{Input}: $DSL$, a DSL to search over \\
\textbf{Input}: $T$, the timeout parameter \\
\textbf{Input}: $M$, the neural network model \\
\textbf{Input}: $K$, number of bootstrapping examples \\
\textbf{Input}: $\tau$, probability threshold \\
\textbf{Output}: correct program if found
\begin{algorithmic}[1]
\STATE \textbf{// Bootstrapping probabilities}
\STATE $prob\_dist \gets $ getProbSpace($M$, $X$, $Y$)
\STATE $counts = $ array of ones of length of $prob\_dist$
\FOR{$k \in \{1 .. K\}$}
\STATE $idx \gets $ Pick a random example index
\STATE $seq \gets $ Pick a random first or first two tokens
\STATE $tmp\_dist \gets $ getProbSpace($M$, $X$, $Y$, $idx$, $seq$)
\STATE $prob\_dist[len(seq):] += tmp\_dist$
\STATE $counts[len(seq):] += 1$
\ENDFOR
\STATE $prob\_dist = prob\_dist / counts$
\STATE
\STATE \textbf{// Enumerate programs from tokens with P $> \tau$}
\STATE $sorted\_progs \gets $ enumerate and sort programs
\STATE
\STATE \textbf{// Evaluate the programs and return the correct one}
\FOR{$\pi \in sorted\_progs$}
\IF{time elapsed $> T$}
\STATE return null
\ENDIF
\STATE $success \gets $ evaluate\_program($\pi$, $X$, $Y$, DSL)
\IF{$success$}
\STATE return $\pi$
\ENDIF
\ENDFOR
\STATE return null
\end{algorithmic}
\end{algorithm}

\begin{algorithm}
\caption{getProbSpace}
\label{alg:getProbSpace}
\textbf{Input}: $X$, input for each example \\
\textbf{Input}: $Y$, target for each example \\
\textbf{Input}: $M$, the neural network model \\
\textbf{Input}: $\tau$, probability threshold \\
\textbf{Input}: $idx$, the example pair index \\
\textbf{Input}: $seq$, the starting seq, empty by default \\
\textbf{Input}: $L$, maximum sequence length \\
\textbf{Output}: Probability distribution over each token in the sequence.
\begin{algorithmic}[1]
\STATE $seq\_len \gets 0$
\STATE $done \gets $ False
\STATE $shiftedSeq \gets [Start Of Sentence] + seq$
\STATE $prob\_dist \gets []$
\WHILE{not done and seq\_len $<$ L}
\STATE probs $\gets$ predict($M$,$X[idx]$,$Y[idx]$,$shiftedSeq$)
\STATE $prob\_dist \gets prob\_dist + [probs]$
\STATE best\_token $\gets$ argmax(probs)
\IF{best\_token is End Of Sentence}
\IF{End Of Sentence is the only class with P $> \tau$}
\STATE $done \gets $ True
\ELSE
\STATE best\_token $\gets$ second most probable class
\ENDIF
\ENDIF
\STATE $shiftedSeq \gets shiftedSeq + [best\_token]$
\STATE $seq\_len \gets seq\_len + 1$
\ENDWHILE
\end{algorithmic}
\end{algorithm}

In the main algorithm (alg. \ref{alg:gridcoder}), once we have estimated our probability space over the DSL, we then proceed to enumerating all the possible programs formed by the token classes whose probability is greater than $\tau$, the probability threshold hyperparameter. If we calculate that the total set of possible programs is greater than a pre-determined number, we gradually increase $\tau$ until the cut-off level results in a manageable number of programs. This is not shown in the pseudo-code, for the sake of brevity. Then, we proceed to sort the programs in decreasing order of their joint probability: we multiply the probabilities associated with all of the tokens in its sequence, truncating at the first \texttt{<End Of Sentence>} token seen.

\subsection{Conditional Independence} 

One aspect of the algorithm that is important, yet is not obvious in the pseudo-code, is the underlying data structure. While typical program induction operates on a tree, GridCoder operates on a flat sequence of nodes. In other words, our probabilities are represented by a list of lists: the probability distribution over a given token is solely dependent on its position in the sequence, rather than being conditional on the previous choices of tokens. That is, to be more exact, it was conditional on the previous (max probability) tokens at the time of auto-regressively calculating the probability distribution. But, at search time, we do not update the choice of selected tokens and requery the model to find the new, conditionally updated probabilities.

Doing this, and maintaing a tree of probabilities instead, would be theoretically correct, because it would respect the natural conditional dependence between primitives. The probability over the next token is theoretically dependent not just on the token's position in the sequence, but on all of the previous selections that were made. However, we choose to make a simplifying assumption of conditional independence between nodes, much like in Bayesian networks. While this simplifying assumption potentially hurts in some cases, overall we believe it yields a significant gain in efficiency: making this assumption means that we only need to call the neural network a few times while bootstrapping the probabilities.

In fact, in the baseline version of this approach, the neural network need only be called exactly once. From there, all of the necessary probability distributions can be acquired. However, it was found empirically that using a bit of bootstrapping helps. Bootstrapping in this context means that we collect the probability distributions as a mean of $K$ decoding iterations (where $K$ is six in our experiments). Each decoding iteration differs in which example pair is provided to the neural network, and which starting token is selected. This bootstrapping idea helps mitigate the rare cases where conditional independence hurts because the neural network outputs an unusually constrained (i.e., falsely confident) set of probabilities on the first inference. Using a few bootstrapped inference trials seems to open up the space of possibilities a bit, introducing a bit of much needed entropy the process.

\subsection{The DSLs}

A few iterations of the DSL were used during the experiments, in which primitives were gradually added to the previous version. They are briefly described here. You can refer to the appendix to see a full list of all primitives in each version, along with a brief description of each primitive.

\paragraph{Version 1} The first version of the DSL contains 74 primitives, which exclusively transform one or two input grids into one grid. These grid-to-grid transformations include things such as changing all pixels of one color to another color, applying grid rotations, mirroring, etc. This choice of primitives was initially made to facilitate research on the LGS paradigm.

\paragraph{Version 2} The second version of the DSL contains 89 primitives. It contains all the primitives in version 1, plus some primitives for object detection and manipulation. In particular, six different versions of object detection were implemented, for various concepts of objectness (what defines an object tends to change from task to task). In addition, it contains new primitives that allowing looping through object lists and applying transforms to them, either in-place, or by cropping all the objects and generating a new grid out of them.

\paragraph{Version 3} The third version of the DSL contains 98 primitives. New primitives were added to allow filtering objects based on a few characteristics such as size and symmetry.

\subsection{VLM Architecture \& Training} \label{sec:architecture}

\begin{figure*}
    \centering
    \includegraphics[scale=0.4]{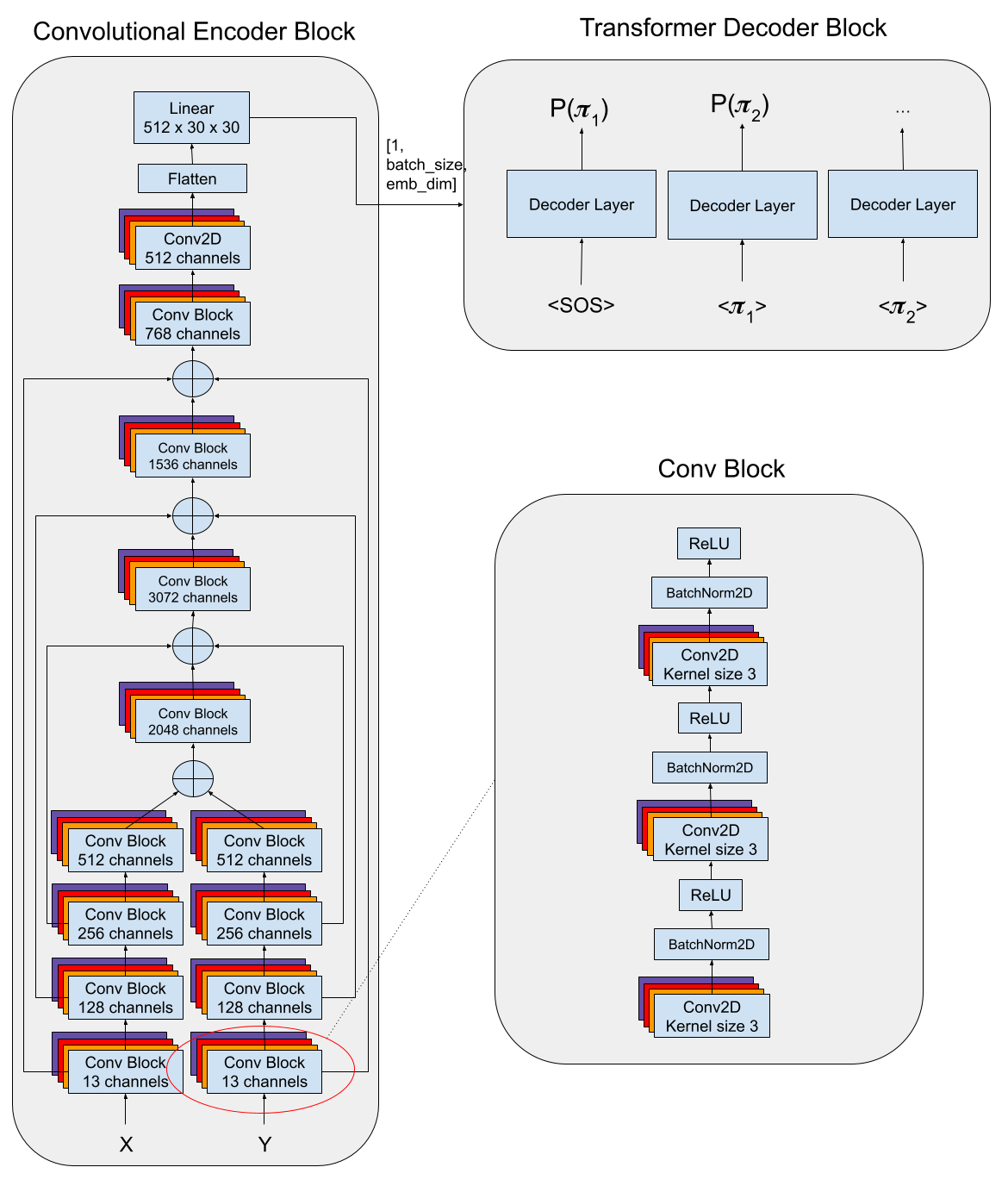}
    \caption{Architecture of VLM}
    \label{fig:VLM}
\end{figure*}

The neural network used to provide the probabilities that guide the search is based on a Vision Language Model (VLM) paradigm (see figure \ref{fig:VLM}). It is a pair of convolutional neural networks, one receiving the input grid, the other the output grid, eventually bridging their intermediate features into one convolutional network. This serves as an encoder, followed by a transformer decoder that processes the target sequence. In our experiments this architecture converged faster and to a higher validation accuracy on our dataset than a \texttt{T5} or \texttt{LongT5} architecture, while allowing us to use a higher parameter count for the same amount of VRAM. Specifically, the model used for the experiments has 452 million parameters, and only one decoder layer with an embedding dimensionality of 512 and 4 attention heads.

Each predicted token sequence forms a program syntax tree in a bottom-up, left-to-right manner. Each token refers to a primitive from the DSL, or to one of the three special tokens: \texttt{<End of Sentence>}, \texttt{<New Level>}, \texttt{<Identity>}. \texttt{<Identity>} can be thought of as a primitive that takes as input a grid and returns that grid without any modification. It is often used as a placeholder in order to disambiguate between different possible interpretations of a program. \texttt{<New Level>} is a special token that indicates that the tokens that follow define the next level in the program syntax tree. The syntax tree is built bottom-up (as the token sequence is read from left to right), and in each level the function argument ordering is defined from left to right in the same order as the token sequence. Here are a few examples to help better understand the syntax structure.

\begin{figure} [h]
    \centering
    \includegraphics[scale=0.5]{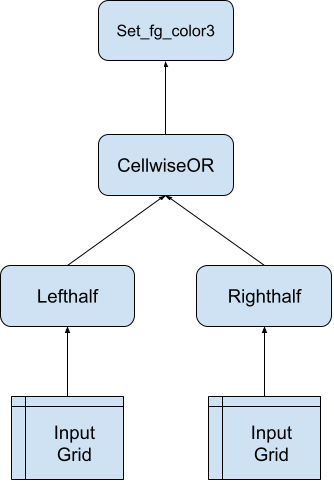}
    \caption{Syntax tree for [\texttt{lefthalf}, \texttt{righthalf}, \texttt{<New Level>}, \texttt{cellwiseOR}, \texttt{<New Level>}, \texttt{set\_fg\_color3}, \texttt{<End of Sentence>}]}
    \label{fig:syntax_tree1}
\end{figure}

\begin{figure} [h]
    \centering
    \includegraphics[scale=0.45]{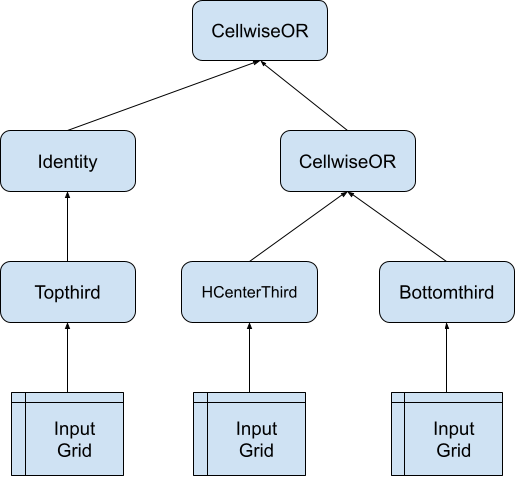}
    \caption{Syntax tree for [\texttt{topthird}, \texttt{hcenterthird}, \texttt{bottomthird}, \texttt{<New Level>}, \texttt{<Identity>}, \texttt{cellwiseOR}, \texttt{<New Level>}, \texttt{cellwiseOR}, \texttt{<End of Sentence>}]}
    \label{fig:syntax_tree2}
\end{figure}

Suppose the following token sequence: [\texttt{lefthalf}, \texttt{righthalf}, \texttt{<New Level>}, \texttt{cellwiseOR}, \texttt{<New Level>}, \texttt{set\_fg\_color3}, \texttt{<End of Sentence>}]. The corresponding syntax tree (see figure \ref{fig:syntax_tree1}) indicates that we take the left half and right half of the original grid separately, merge the pixels using OR logic (i.e., if any pixel is a foreground pixel, we place a foreground pixel of the same color on the target grid, prioritizing the color of the first argument if both are foreground pixels), and then change all of the non-zero pixels to color three.

A more advanced example is the following token sequence: [\texttt{topthird}, \texttt{hcenterthird}, \texttt{bottomthird}, \texttt{<New Level>}, \texttt{<Identity>}, \texttt{cellwiseOR}, \texttt{<New Level>}, \texttt{cellwiseOR}, \texttt{<End of Sentence>}]. Here the corresponding syntax tree (figure \ref{fig:syntax_tree2}) forms a program that splits the original grid into three sub-grids vertically, and then merges them with OR logic. Specifically, in the second level it passes on the output of \texttt{topthird} intact to the upper level, which means that it first merges the output of \texttt{hcenterthird} and \texttt{bottomthird}. Finally, at the last level, it merges \texttt{topthird} with the output from the previous level that merged \texttt{hcenterthird} and \texttt{bottomthird}.

The training was first done on the DSL version 1 data, until convergence. Then, the DSL version 2 and DSL version 3 models were separately fine-tuned on their respective datasets, starting from the pre-trained DSL version 1 model. The DSL 2 and DSL 3 datasets contain training examples from the previous versions as well (not just exclusively tasks related to the newly added primitives). The learning rate selected was 0.0001, and the weight decay is 0.005. For DSL version 1, a total of around 5M training samples (i.e., input-output grid pairs + ground-truth program sequence for distinct tasks) were generated. For DSL version 2, around 600K training samples were generated, and around 300K for DSL version 3 (because data generation is slower). More data would have been preferable, but time was a significant constraint.

\subsection{Training Data Generation}

The training data for the VLM is the output of procedurally generated tasks. The concept is to identify task "meta-patterns" or categories in the ARC-AGI Training Set, and to implement data generators that reproduce them, while randomly varying several aspects of the tasks. Different DSL versions have different task generators associated with them.

\paragraph{DSL version 1} In this version, there are only three task generators implemented. The input grids used to generate the output grids are randomly sampled from two sources: manually pre-generated grids (mostly drawn from the ARC-AGI Training set), and randomly generated grids. Input grids that match grids found in the ARC-AGI Evaluation set are not allowed, though it is not impossible for the randomly generated grids, by extreme luck, to reproduce grids similar to those in the ARC-AGI Evaluation set.

The first task generator consists of randomly generating trivial tasks that are either made up of one primitive (that takes as input a grid and generates a grid as outupt), or of two primitives composed together. The two primitive tasks are not entirely random: a script that generates all two-primitive permutations of the DSL was created, but several heuristics were used to prevent non-sensical or redundant compositions. Overall, this task generator has $796$ distinct tasks.

The second task generator is based on a "split-merge" pattern that was observed. This task category produces tasks in which the goal is to split the input grid in some way, and then merge them back together into one grid based on \textit{OR, XOR, AND, NOR} or \textit{Difference} logic (see the appendix for details). Finally, one last global grid-to-grid transform can be randomly applied on the resulting grid. The possible split patterns are two-way to four-way horizontally or vertically, or four-way by taking each quadrant separately. These various parameters (split method, logical operator, optional last transform) are randomized and combined to generate a whole range of possible tasks that follow the same "meta-pattern".

The third task generator is based on a "tiling" pattern. In this task category, the idea is to apply various rotational and mirroring transforms to the input grid, and concatenate its transformed variants together in some way to produce a larger output grid. The possible individual transforms are randomly picked from rotations, the identity function, or flipping the grid horizontally or vertically. The tilling patterns that can be generated are 2x2, 3x3, horizontal concatenation of 2 to 4 grids, and vertical concatenation of 2 to 4 grids.

\paragraph{DSL version 2} This DSL version includes a new general-purpose task generator whose pattern is to randomly pick a number of objects to generate, to randomly pick some transforms to apply to them (including no transform at all as a possibility), and to decide whether to apply those transforms to the objects in-place, or to crop the objects and tile them together into some pattern (which itself is randomly determined among a few possible templates). Finally, there is an optional post-processing transform, applied to the grid as a whole, that can be selected.

\paragraph{DSL version 3} This DSL version includes three new task generators that are a bit more niche than the one added in DSL version 2. The first one, the object selector task generator, consists of generating random objects and deciding randomly whether the task is to keep one, or to filter out some objects based on one of the following characteristics: horizontal symmetry or lack thereof, vertical symmetry or lack thereof, object size (number of foreground pixels it contains), and the number of sub-objects it contains. As for most other task generators, a random post-processing transform can be applied to the output grid.

The windowing task generator produces frame-like objects, i.e. non-filled rectangles of uniform color that can contain randomly generated pixels or objects. The objective of the task is to apply randomly selected transforms to the inside of the frames (with or without the rectangular boundary) and then either crop the object, if there is only one, or apply those transforms in place.

Finally, the object recombiner task generator creates random objects on a grid, decides whether to split the objects horizontally or vertically, and applies some color-based modification to one of the halves. It then recombines the halves into whole objects.

\section{Learning the Transform Space}

This approach can be thought of as an evolution of LPS, while bringing back the execution-guided aspect of LGS. The concept is to train a model such that, given an intermediate or starting program state, and a target grid, it predicts the probability distribution over the DSL for the next token. In other words, the main difference with LPS is that we explicitly feed back into each decoding step some notion of the intermediate state of the program. Thus, the algorithm can leverage the efficiency of auto-regressive decoding with the ability to evaluate the transformations in and of themselves regardless of whether this program pattern has been seen during training.

The emphasis is on learning what transformations are required to bring the program execution from its current state to the target grid. Initially, this was the intended result of the GridCoder algorithm, however further analysis reveals that the latter does not maintain such an hidden state implicitly, it only superficially relies on the encoding of the input-output grid and the token sequence generated so far. See the Discussion section for a deeper analysis.

LTS borrows the same program syntax and auto-regressive sequence supervision as in GridCoder. In a more basic implementation, the intermediate program state could simply be the last full intermediate grid produced by the program execution. The experiments associated with Table \ref{tab:transformation_space} make this basic assumption. In a more sophisticated version, however, there needs to be a hidden latent state maintained throughout the decoding process that gets updated at each step.

The challenging aspect of maintaining this generalized intermediate state is that the output of a primitive can be anything from a Grid to an integer, a Boolean, a list of integers, a list of Grids, etc. A clever mechanism must be designed to allow embedding into a fixed vector space this dynamic and diverse range of outputs, which is left to future work.

\section{Experiments \& Results} 

This section presents the results of three experiments. First, a comparison of the performance of different approaches on the ARC evaluation set is made. Second, the increments of performance of the selected approach (GridCoder) are shown, as new primitives are added to the DSL. Finally, an experiment on structurally out-of-distribution tasks is shown, to illustrate the limitations of the selected approach and the promising capabilities of the suggested new approach (\textit{Learning the transformation space}). 

\subsection{Performance Comparison on ARC-AGI Eval Set}

Table \ref{tab:overall} reports the success rate (as a percentage of solved tasks) of various approaches on the ARC-AGI evaluation set. We show the numbers in terms of the absolute ARC-AGI evaluation set performance, but also in terms of the subset of the ARC-AGI evaluation set that is theoretically solvable according to the DSL. DSL Version 1 is used for these experiments (see section titled "The DSLs"). This subset refers to the 29 tasks ($7.25\%$ of full evaluation set) that can theoretically be solvable in the current DSL. All the other tasks are impossible to solve regardless of how efficient the search is.

For each task, a time budget of up to 15 minutes to find the answer is allowed. For \textit{GridCoder} and \textit{GridCoder cond.}, the 5-minute time budget performance is also shown, in order to better differentiate the impact of the conditional independence assumption. The following algorithms are compared in this experiment:

\paragraph{GridCoder} This is the selected approach for the associated ARC-AGI submission, elaborated upon in the \textit{Learning the program space} section.
\paragraph{GridCoder cond.} This is a variant of GridCoder where the conditional independence assumption is not made. The objective is to show that the conditional independence assumption is an improvement.
\paragraph{VLM-only} This is the pre-trained neural network used in GridCoder, but without the search component. The objective is to show the necessity of search.
\paragraph{MCTS+VLM} This uses \textit{Monte-Carlo Tree Search}, instead of GridCoder, for the search algorithm. Specifically, it uses a MuZero-like approach \cite{muzero} where the probabilities instruct the primitive selection at each node, and a pixelwise similarity metric (same as in \textit{Pixelwise sim}) is used to backpropagate the value.
\paragraph{Pixelwise sim} This is an LGS approach. Here, \textit{pixelwise} refers to a simple hand-crafted pixelwise comparison of the grid, and the percentage of identical pixels is the similarity heuristic. There is no learning in this version.
\paragraph{Learned sim} Similar to \textit{Pixelwise sim}, but the heuristic is a pre-trained Transformer encoder that outputs a similarity heuristic between 0 and 1, as discussed in the \textit{Learning the Grid Space} section.

\begin{figure}
    \centering
    \includegraphics[scale=0.3]{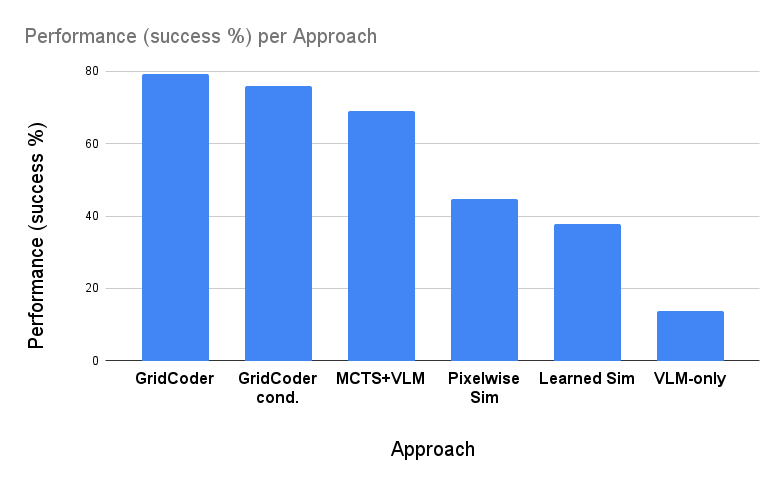}
    \caption{Success rate on the 29-task solvable set from ARC-AGI evaluation set}
    \label{fig:exp1_results}
\end{figure}

\begin{table*}
\begin{center}
\begin{tabular}{|c | c | c | c |} 
 \hline
 Algorithm & ARC-AGI Eval Set \% (15m) & Solvable subset \% (15m) & Solvable subset \% (5m) \\
 \hline\hline
 \textbf{GridCoder} & \textbf{5.75} & \textbf{79.3} & \textbf{79.3} \\ 
 \hline
 GridCoder cond. & 5.5 & 75.9 & 72.4 \\
 \hline
 MCTS+LVM & 5 & 69 & -\\
 \hline
 Pixelwise Sim & 3.25 & 44.8 & - \\
 \hline
 Learned Sim & 2.75 & 37.9& - \\
 \hline
 VLM-only & 1 & 13.8 & - \\
 \hline
\end{tabular}
\caption{Success rate on full ARC-AGI evaluation set at 15-minute budget, on subset of 29 solvable tasks at 15-minute budget, and on subset of 29 solvable tasks at 5-minute budget.}
\label{tab:overall}
\end{center}
\end{table*}

It should be noted that the two LGS implementations tend to fail on tasks whose ground truth program has more than 5 or 6 primitives, explaining its relatively low performance. It captures the low-hanging fruit of relatively small programs quite well, but it is not efficient enough to solve the programs that have longer descriptions.

\subsection{GridCoder on Different DSLs}

These results make full use of the three presented versions of DSLs. The objective is to study the scaling characteristics of the LPS approach and determine whether scaling this proof-of-concept to a competitive solution is plausible. This is done by training three separate VLM instances, one on each of the three DSLs and their associated training data generators. This can be thought of as a curriculum progression, where the DSL version 3 model includes all primitives and task generators of DSL version 2, which itself includes all primitives and task generators of DSL version 1.

The performance (success rate) on the ARC-AGI Evaluation set  can be found in Table \ref{tab:DSL_perf}. For each, the performance on the DSL version 1 solvable tasks (a subset of 29 tasks from the ARC Evaluation Set), with a 5-minute time budget, is used to show whether there is regression on previous tasks as we grow the DSL. For the sake of transparency, the hidden test set performance is also displayed. 

The inference times per task are presented in Table \ref{tab:DSL_times}, to determine whether the solution time scaling with respect to DSL size is super-linear, linear, or sub-linear. The latter is a more attractive quality, since it suggests that the DSL can be grown to solve more tasks, while the former would suggest a solution that scales unfavorably. The average inference time per task, per DSL, and its standard deviations are reported. Additionally, the normalized time per DSL size, i.e. the number of seconds per task on average divided by the number of primitives in the DSL, is shown in order to directly observe whether we have such a super-linear or sub-linear progression.

\begin{table*}
\begin{center}
\begin{tabular}{| c | c | c | c |} 
 \hline
 DSL Version & ARC-AGI Eval Set \% & DSL 1 subset \% & Hidden test set \% \\
 \hline\hline
 Version 1 & 5.75 & 79.3 & 0\\ 
 \hline
 Version 2 & 7.75 & 79.3 & 1 \\
 \hline
 \textbf{Version 3} & \textbf{8.25} & \textbf{79.3} & \textbf{1}\\
 \hline
\end{tabular}
\caption{Success rates of different DSL versions on full ARC-AGI evaluation set at 5-minute budget, and on the 29-task DSL version 1 subset at 5-minute time budget. ARC-AGI hidden test set performance is also presented.}
\label{tab:DSL_perf}
\end{center}
\end{table*}

\begin{table}
\begin{center}
\begin{tabular}{| c | c | c | c |} 
 \hline
 DSL Version & Mean per task & Mean per primitive \\
 \hline\hline
 Version 1 & 4.74 (1.07) & 0.064 \\ 
 \hline
 Version 2 & 5.82 (1.22) & 0.064 \\
 \hline
 Version 3 & 5.66 (1.22) & 0.057 \\
 \hline
\end{tabular}
\caption{Mean number of seconds taken to solve a task, by DSL. Standard devations in parentheses. Mean number of seconds taken to solve a task, normalized by number of primitives in a DSL.}
\label{tab:DSL_times}
\end{center}
\end{table}

\subsection{Generalization of LTS}

The purpose of this experiment is to suggest a promising new method that mitigates the limitations of the selected GridCoder approach (see Discussion section), by indicating preliminary empirical results on its out-of-distribution generalization characteristics. 10 new tasks are hand-crafted, specifically selected to guarantee that there is no structurally similar task ever generated in the training data. One of them is picked from the ARC-AGI evaluation set, task \#48131b3c, since it has been observed that GridCoder fails to generalize to this task.

A high-level description of the 10 tasks follows:
\begin{enumerate}
    \item Task \#48131b3c from ARC-AGI evaluation set: tile the original grid 2x2, and invert the colors. The training data generator never presents the post-tiling inversion of colors.
    \item Hand-crafted task 1: gravitate the pixels to the left, then gravitate the pixels upward, then change the foreground pixels' color to aquamarine. The training data generator never generates tasks that involves the application of three primitives in a row (or longer). The maximum is two.
    \item Hand-crafted task 2: rotate the grid 90 degrees, upscale it horizontally by two, and then upscale it vertically by two.
    \item Hand-crafted task 3: same as task 2, but add an extra operation of horizontal mirroring at the end.
    \item Hand-crafted task 4: same as task 3, but add an extra operation of color inversion at the end.
    \item Hand-crafted task 5: tile the original grid 2x4 while alternating 180-degree rotation with the identity transform on each tile. This is out-of-distribution because a 2x4 tiling is never generated in the training data. The closest are either 2x2 or 1x4.
    \item Hand-crafted task 6: tile the smallest object in the grid twice horizontally (i.e. concatenate it with itself horizontally), and use that as output grid. In the training data, tiling tasks and object selection tasks are never combined.
    \item Hand-crafted task 7: crop the the object that contains the largest number of sub-objects, rotate it 90 degrees, and then duplicate the top row and the bottom row. The training data only at most applies 1 post-processing transform on these types of object cropping tasks.
    \item Hand-crafted task 8: filter out the largest object in the grid and then rotate the grid 270 degrees. The training data generation never applies rotation primitives to the output of an object filtering task.
    \item Hand-crafted task 9: crop the largest object in the grid, split it in half horiziontally, and merge the left and right halves with cellwise OR logic. In the training data, object cropping and "split and merge" types of tasks are never combined.
\end{enumerate}

The experiment consists of simulating a model that was trained to decode in an execution-guided way (see the section \textit{Learning the Transformation Space}), receiving execution feedback at each step along the way. While the potential impact on inference time is not simulated, the task attempt is considered a success if it is possible to reach the solution by re-launching the current search algorithm on the intermediate output of one of the program sequences that get evaluated in the previous run. As a result, the program is expected to provide a probability over the program space from a starting point that is the intermediate output of a previous program -- hence it is given the opportunity to compose sub-programs.

This proxy experiment was done, instead of correctly training a model that learns to decode in such a way, due to lack of time. It is intended as an approximation, or suggestion of what is potentially achievable, if we fully train the model to receive an execution output at each decoding step. Table \ref{tab:transformation_space} indicates the results on the 10 tasks, comparing the non-execution-guided GridCoder to the execution-guided GridCoder.

\begin{table}
\begin{center}
\begin{tabular}{| c | c | c |} 
 \hline
 Task & GridCoder & Execution-guided \\
 \hline\hline
 Task \#48131b3c & NO & YES \\ 
 \hline
 Task 1 & YES & YES \\ 
 \hline
 Task 2 & NO & YES \\ 
 \hline
 Task 3 & NO & YES \\ 
 \hline
 Task 4 & NO & YES \\ 
 \hline
 Task 5 & NO & NO \\ 
 \hline
 Task 6 & NO & NO \\ 
 \hline
 Task 7 & NO & YES \\ 
 \hline
 Task 8 & NO & YES \\ 
 \hline
 Task 9 & NO & YES \\ 
 \hline
 \textbf{Total (\%)} & \textbf{10} & \textbf{80} \\
 \hline
\end{tabular}
\caption{Success on out-of-distribution tasks, comparing the selected GridCoder with the suggested execution-guided search.}
\label{tab:transformation_space}
\end{center}
\end{table}

Task 5 fails on both approaches because when the neural network sees a large target grid (of 2x4 in this case), instead of attempting a 2x2 solution or a 1x4 solution which can then be scaled by composition to the correct 2x4 solution, it goes directly for 3x3. So the partial solutions that it suggests cannot be used to produce the correct solution.

Task 6 fails on both approaches because it fails to see what needs to be done after the initial object cropping. It attributes near-zero probabilities to the primitives that yield the required horizontal tiling. Instead it seems to suggest programs that would mirror the input, horizontally or vertically.

\section{Discussion}

\subsection{Generalization Characteristics}

In the first experiment, the gap between \textit{VLM-only} and \textit{GridCoder} is strictly an out-of-distribution generalization gap. The fact that the search-enabled algorithms obtain a significantly better performance than the \textit{VLM-only} approach suggests that searching over the probability space significantly extends the neural network's reach beyond its training distribution. However, generalization obstacles remain, as indicated by the experiments in out-of-distribution generalization where the GridCoder approach only succeeds in 1 out of the 10 tasks. Furthermore, we note the following additional GridCoder failure cases:

\begin{figure}
    \centering
    \includegraphics[scale=0.35]{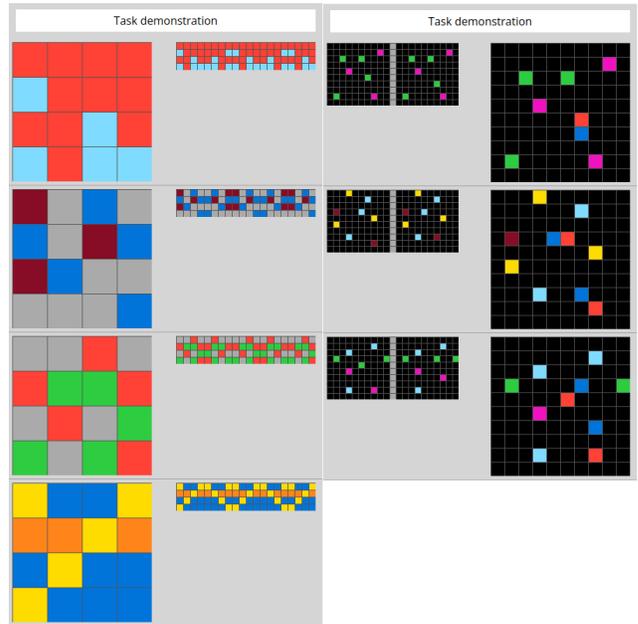}
    \caption{Two examples of GridCoder failures. Left: task bc4146bd. Right: task d47aa2ff.}
    \label{fig:failure_examples}
\end{figure}

\paragraph{bc4146bd.json} This task (see figure \ref{fig:failure_examples}) consists of "tiling" the original grid fives time horizontally, using various grid transformations (e.g. rotations) each time. While the neural network was exposed to conceptually similar tasks during training, it has never seen fivefold horizontal (or vertical) tiling at all. As such, it struggled to generate the program structure that allows this fivefold tiling. 

\paragraph{d47aa2ff.json} This task (see figure \ref{fig:failure_examples}) consists of copying over to the output grid the pixels that are common to both halves of the input grid. However, when there is a discrepancy between both halves, if the pixel only appears on the left grid, it must be colored red in the output, and if it is the reverse it must be colored blue. While the DSL allows us to solve this task in principle, the solution is quite elaborate and there is no conceptually similar task in the training data.

The vast improvement in generalization capabilities from \textit{VLM-only} to \textit{GridCoder} can appear to be in opposition to the poor structural generalization findings from Table \ref{tab:transformation_space}. We hypothesize that the out-of-distribution uncertainty that gets resolved by the probabilistic search is mainly related to a grid-level distribution shift. The test grids are distinct from the training ones, hence it struggles with properly predicting the types of per-grid transformations. This results in a relatively high entropy when choosing a primitive that produces some kind of grid-to-grid transformation. 

However, we observe a very low entropy on tokens that set the overall program structure (for example, when differentiating "tiling" types of tasks from "split and merge" types of tasks). As a result, the search is able to resolve grid-level distribution shift by searching over the relatively high-entropy distribution, while the very low entropy distribution of program structure across task categories means that it cannot generalize out-of-distribution in that sense.

An example will better illustrate this. Consider, for example, the predicted probabilities for task \texttt{59341089.json}: 

\paragraph{Probabilities at position 0:}
\begin{itemize}
    \item \texttt{<Identity>}: 0.56
    \item \texttt{vmirror}: 0.10
    \item \texttt{hmirror}: 0.17
    \item \texttt{rot90}: 0.02
    \item \texttt{rot180}: 0.11
    \item \texttt{rot270}: 0.03    
\end{itemize}

\paragraph{Probabilities at position 1:}
\begin{itemize}
    \item \texttt{<Identity>}: 0.43
    \item \texttt{vmirror}: 0.13
    \item \texttt{hmirror}: 0.22
    \item \texttt{rot90}: 0.02
    \item \texttt{rot180}: 0.15
    \item \texttt{rot270}: 0.05    
\end{itemize}

\paragraph{Probabilities at position 2:}
\begin{itemize}
    \item \texttt{<Identity>}: 0.24
    \item \texttt{vmirror}: 0.16
    \item \texttt{hmirror}: 0.33
    \item \texttt{rot90}: 0.02
    \item \texttt{rot180}: 0.20
    \item \texttt{rot270}: 0.04    
\end{itemize}

\paragraph{Probabilities at position 3:}
\begin{itemize}
    \item \texttt{<New Level>}: 0.02
    \item \texttt{<Identity>}: 0.31
    \item \texttt{vmirror}: 0.14
    \item \texttt{hmirror}: 0.26
    \item \texttt{rot90}: 0.03
    \item \texttt{rot180}: 0.19
    \item \texttt{rot270}: 0.04    
\end{itemize}

\paragraph{Probabilities at position 4:}
\begin{itemize}
    \item \texttt{<New Level>}: 0.99
\end{itemize}

\paragraph{Probabilities at position 5:}
\begin{itemize}
    \item \texttt{hconcat}: 1.00
\end{itemize}

\paragraph{Probabilities at position 6:}
\begin{itemize}
    \item \texttt{hconcat}: 0.99
\end{itemize}

\paragraph{Probabilities at position 7:}
\begin{itemize}
    \item \texttt{<New Level>}: 0.99
\end{itemize}

\paragraph{Probabilities at position 8:}
\begin{itemize}
    \item \texttt{hconcat}: 0.99
\end{itemize}

\paragraph{Probabilities at position 9:}
\begin{itemize}
    \item \texttt{<End of Sentence>}: 0.99
    \item \texttt{hconcat}: 0.01
\end{itemize}

Only the classes whose probability is 0.01 or greater are shown. This task consists of tiling the input grid four times horizontally, while applying the following transforms to the grid, in order, from left to right: [\texttt{hmirror}, \texttt{<Identity>}, \texttt{hmirror}, \texttt{<Identity>}]. The token positions responsible for the "skeleton", or "structure" of the 4-way tiling program have very high probabilities (low entropy) associated to them. These are the token positions 4 to 9. The token positions responsible to determining what transformations are applied to the grid, and in what order, are, however, high entropy.

Further training the model will not solve these generalization issues. What has been observed is that the solved tasks are solved more rapidly when the neural network is better trained, because the probabilities are more accurate and less noisy. However, a token sub-sequence that has never been observed during training will still retain a probablity near zero (obviously, since the probability of that sub-sequence in the training set is exactly zero), and a token sub-sequence that is always observed under a given "type" of input-output grid pairs will have a probability of one, since nothing else has ever been observed during training. This is precisely the behaviour that limits the generalization capabilities of LPS methods such as GridCoder.

Adding a bit of entropy such that no probability is ever zero or one, could help break through the hard probability barriers, at the expense of a vastly increased search space. Furthermore, this is unlikely to allow the search to discover extremely low probability (high complexity) modifications such as the one required to go from the most similar program for task bc4146bd (from figure \ref{fig:failure_examples}) to its correct solution.

This is where Learning the transformation space  (adding an execution-guided feedback and re-evaluating the probabilities on the intermediate grids) can help break through this generalization barrier, as suggested by the experimental results of Table \ref{tab:transformation_space}.

\subsection{Scaling the DSL}

A potential criticism for the LPS approach is that, as we grow the DSL, the search space might become exponentially more complex, so the solution time slows down proportinately. This would make this approach unlikely to scale to a competitive solution on ARC-AGI. Another possibility is that, as we add new primitives to the DSL, and train on new tasks, performance degrades on previous tasks.

Tables \ref{tab:DSL_perf} and \ref{tab:DSL_times} paint a positive picture of the scaling properties of the LPS approach. First, we see from Table \ref{tab:DSL_perf} that as we grow the DSL from version 1 to version 3, the tasks that the approach was able to solve in DSL version 1 are still solved in subsequent sections. There is no forgetting, and no significant loss in search performance that degrades the success rate. Meanwhile, the overall performance on the ARC-AGI evaluation set increases as we add primitives and corresponding training tasks.

Furthermore, we see that from Version 1 to Version 3, the mean solution time per primitive does not increase. In fact, it seems to decrease a bit from Version 2 to Version 3. This may simply be a fluke from having a better trained model in Version 3 and Version 2. As a reminder, DSL Version 1 has 74 primitives, DSL Version 2 has 89 primitives, and DSL Version 3 has 98 primitives. In summary, the solution time scales sub-linearly as we add more primitives to the DSL, which is a desirable property.

\subsection{Flaws in LGS}

The LGS approach reveals itself to be somewhat inefficient, though \textit{a priori} it may be better at structural out-of-distribution generalization (because it is not trained on specific task solutions). Extensive work has been done to understand the causes of this inefficiency. The main hypothesized causes are as follows:

\paragraph{A* versus Q*} The idea of a model that takes as input two grids and predicts their similarity suggests an A*-style usage pattern:

\begin{enumerate}
    \item Each transformation in the DSL is applied to the input grid to obtain some intermediate grid.
    \item Each pair of intermediate grid and target grid is fed to the similarity model to obtain their respective similarities.
    \item The primitive that yields the biggest increase in similarity is selected.
    \item This is done iteratively until a solution is found or some budget limit is reached.
\end{enumerate}

\citet{q_search} show that this approach is both temporally and spatially inefficient, and propose Q* search, which can be differentiated as follows:

\begin{enumerate}
    \item The input grid and output grid pair is fed to the neural network, which in one shot simultaneously predicts all of the similarity (or \textit{cost-to-go}) values for each of the primitives.
    \item The best one is selected, and used to generate the next intermediate state.
    \item This is done iteratively until a solution is found or some budget limit is reached.
\end{enumerate}

Hence, where the grid-to-grid similarity approach results in N evaluations where N is the size of the DSL, the Q* approach has a constant number of evaluations with respect to the DSL size. Note that this means we are inherently moving from a LGS approach to a LTS approach. Instead of having a model that learns a latent grid space contrastively and outputs similarities between two vectors, we have a model that necessarily has to learn the effect of applying each transformation from the DSL, and evaluate its cost-to-go. This becomes even more evident when we supervise this model to output probabilities over the DSL instead of a cost-to-go.

The choice of \textit{cost-to-go} or probability in this context is debatable, but we hypothesize that learning probabilities is more direct and efficient than the cost-to-go. This is the case both from the standpoint of learning the underlying relationships and from the standpoint of generating training data. It is easy to accidentally produce incorrect cost-to-go labels from procedurally generated programs with redundant primitives.

\paragraph{Incomplete information} A program is a syntax tree, not a flat sequence of grid-to-grid operations. Because of that, as we execute a program from the bottom of the tree upwards, there are potentially multiple operations that need to be done in "parallel" (not computationally speaking, but syntactically speaking). Thus, at any given point in execution time, the intermediate state of a program is in fact a list of objects (not necessarily grids, it can be lists of indices, integers, etc.).

\begin{figure}
    \centering
    \includegraphics[scale=0.35]{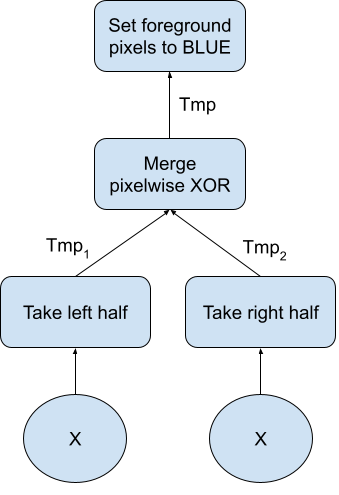}
    \caption{Example of a task solution}
    \label{fig:incomplete_information}
\end{figure}

To select the next primitive based only on the output of one of the previous transformations means that we rely on incomplete information. As an example, refer to figure  \ref{fig:incomplete_information}. In this diagram, $X$ represents the input grid. The left half and right half are taken, and merged together using XOR logic. Consider what happens at training time, when the model is learning to predict the cost-to-go for the first primitive (i.e.  "Take left half" or "Take right half").

The choice of whether to use XOR, OR, AND, NOR, or any other merge-based logic requires knowing which two input grids are fed to these primitives. There are several different cost-to-go values that correspond to the same pair (output of "Take right half" and the target grid), depending on the left half that is used in the procedurally generated program. Because the model necessarily involves a similarity between exactly two grids, the model has to learn without knowing the output of the left half operation. Thus, the model is trained on incomplete information, resulting in weak cost-to-go estimates.

\paragraph{ARC-AGI is not a grid-to-grid problem} ARC-AGI is fundamentally not a grid-to-grid problem, when we consider the execution trace of a solution. That is, most of the steps involved in solving an ARC-AGI task are not strictly grid-to-grid transformations and thus do not result in an intermediate grid that can be evaluated for its similarity to the target. This is related to the previous argument, in the sense that a solution program for most ARC-AGI tasks is not a simple sequence of grid-to-grid transformations, but instead can be thought of as a sequence of intermediate program states which themselves are lists of various kinds of intermediate objects and values. As such, the grid-to-grid similarity concept only potentially allows guiding a small portion of the overall search for a solution.

In summary, these conceptual flaws, combined with the inferior empirical results, led to the choice of selecting the LPS approach instead. However, it should be noted that the LGS idea also presents some essential characteristics that the selected LPS approach is lacking. This is why we attempt to combine the best of both approaches via the suggested LTS approach.

\subsection{Limitations and Future work}

\paragraph{Limited DSL} The work done so far has been a proof-of-concept centered on finding an optimal neurally-guided search algorithm given some pre-determined DSL and a set of tasks that are known to be solvable within this DSL. Very limited work has been done on improving the DSL. This is, therefore, the main limitation of this work. 

Scaling the algorithms presented here to a competitive level will certainly involve completing the DSL by introducing all of the necessary primitives required to solve all ARC training and evaluation tasks. This is not as simple as going through the existing tasks and implementing primitives for every high-level functionality seen. 

We aspire to making the DSL more flexible and general as well, as we currently deem it too high-level and specific. We aim for a certain compromise between a purely generic programming language and a highly specific DSL such as the current one. We expect this work, on its own, to make our approach competitive on the ARC hidden test set.

\paragraph{Functional DSL} Based on the work by \citet{relational_DSL}, it appears that a relational type of DSL can lead to more efficient search than the kind of functional paradigm currently being used in these experiments. Intuitively, relational decomposition of tasks means that partial verification of a subset of the rules is possible while searching for a solution. In contrast, our current approaches can only verify the full program in a binary "fully correct" or "fully incorrect" manner. Relational decomposition therefore allows a more gradual progression towards the full solution.

\paragraph{\textit{get\_objects} primitives} In retrospect, the current implementation of six different primitives for object detection is very sub-optimal. In spite of having implemented six different notions of what is an object, there are still many cases where object detection primitives fail. Going forward, we will instead implement this primitive as an "object segmentation" neural network that, we hope, will allow us to reduce the number of object detection primitives to one, while simultaneously covering more use cases.

\paragraph{Program syntax} The program representation syntax needs to be improved. First, the difference in program structure between four tiles and five tiles is relatively large, when in principle it should be quite small. Here is an example program sequence that tiles an input grid four times horizontally:

[\texttt{hmirror}, \texttt{<Identity>}, \texttt{hmirror}, \texttt{<Identity>}, \texttt{<New Level>}, \texttt{hconcat}, \texttt{hconcat}, \texttt{<New Level>}, \texttt{hconcat}, \texttt{<End of Sentence>}]

Here is the program sequence to tile an input grid five times horizontally:

[\texttt{hmirror}, \texttt{<Identity>}, \texttt{hmirror}, \texttt{<Identity>}, \texttt{hmirror}, \texttt{<New Level>}, \texttt{hconcat}, \texttt{hconcat}, \texttt{<Identity>}, \texttt{<New Level>}, \texttt{hconcat}, \texttt{<Identity>}, \texttt{<New Level>}, \texttt{hconcat}, \texttt{<End of Sentence>}]

There is a difference of five tokens between the two, spread throughout all of the levels of the program. Clearly, the conceptual similarity of these two tasks is not reflected in the syntactical similarity of their solutions. This certainly does not help the search. 

Another problem is that a lot of duplication is necessary in the lower levels. For example, in the last task used for the results of Table \ref{tab:transformation_space}, the goal is to crop the largest object, split it in half horizontally and merge it pixelwise using OR logic. In order to take the left half and the right half on the same input, the current program syntax forces us to duplicate the entire sub-tree that generates the cropped object. The corresponding solution program is:

[\texttt{get\_objects1}, \texttt{get\_objects1}, \texttt{get\_object\_size}, \texttt{get\_objects1}, \texttt{get\_objects1}, \texttt{get\_object\_size}, \texttt{<New Level>}, \texttt{<Identity>}, \texttt{for\_each}, \texttt{<Identity>}, \texttt{for\_each}, \texttt{<New Level>}, \texttt{keep\_largest}, \texttt{keep\_largest}, \texttt{<New Level>}, \texttt{lefthalf}, \texttt{righthalf}, \texttt{<New Level>}, \texttt{cellwiseOR}, \texttt{<End of Sentence>}]

The entire sub-tree [\texttt{get\_objects1}, \texttt{get\_objects1}, \texttt{get\_object\_size}, \texttt{<New Level>}, \texttt{<Identity>}, \texttt{for\_each}, \texttt{<New Level>}, \texttt{keep\_largest}] needs to be duplicated twice, thereby increasing the solution by six tokens. This is obviously not very efficient.

\paragraph{Learning the transform space} The empirical results suggest that GridCoder is limited to proposing program structures that have been seen during training. The neural network does not get to observe explicitly or even generate implicitly the intermediate state of a program as it derives its solution. Instead, it currently has to be able to solve all of it in one go based on the provided input-output examples. 

In other words, as it generates new tokens in the decoding loop, it does not get to look at the intermediate program state to decide on the next steps.  We certainly do not do this explicitly (yet), and we know from how transformer decoders work that it does not naturally maintain a hidden state between decoding steps. It only gets to look at the decoded sequence so far, as well as the output of the encoder. As a result, it tends to lazily rely on positional encoding and predetermined sequence templates that it learned during training. The consequence of this is that, as the search stumbles into a territory where the model should know what to do to transform the intermediate grids into the target grids, it cannot leverage this knowledge. 

It could be argued that a sufficiently flexible transformer decoder module could potentially learn to maintain implicitly the current intermediate state by applying the current token sequence to the encoder output. This is, in fact, what we were hoping to see, and our architectural choice of a particularly lean decoder could be the culprit. However, it's not clear what would encourage it to learn to do this if it can lazily rely on the positional encoding and the token sequence as it does at the moment. Even if this were to be true, however, it is still the case that explicitly feeding back the intermediate state would constitute an inductive bias that can guide the learning process towards faster and better convergence. This makes LTS a promising paradigm regardless.

\section{Related Work}

\paragraph{Execution-guided program induction} Execution-guided program induction can apply to both the LGS and LTS paradigms, in our current nomenclature. Much work has focused on this approach \cite{egs1, egs2, egs3}. Recently, \citet{tree_diffusion} use an image similarity heuristic to discover programs that transform an input image into a target image. The training procedure follows a denoising paradigm, and the search process is based on a diffusion, in which program edits are gradually made to arrive at the denoised image. 

\paragraph{Neurally-guided probabilistic program induction} DreamCoder \cite{dreamcoder} is a program induction approach not unlike the one proposed here, under what we refer to as the LPS paradigm. There is a neural network that learns to output probabilities over the DSL, and a search algorithm suggests programs to evaluate against the example inputs and outputs. It brings as its main innovations a DSL growth mechanism, in which frequently occurring semantically equivalent subroutines are added to the DSL. 

GridCoder does not use such a DSL growth mechanism. On one hand, we see the limited amount of natural ARC-AGI examples as an obstacle to making this feature particularly useful. Correctly fine-tuning the granularity of the DSL, and having the relevant reusable primitives, is something we prefer to do manually. Similarly, DreamCoder has a "dreaming" functionality in which new tasks are randomly generated from the DSL in order to train the neural network. We found that a purely random task generation approach tends to generate several incoherent, redundant, unnatural tasks. Instead, in GridCoder we adopt an intermediate position in which we manually design high-level task categories or concepts, but allow some random variation in their parameters and details.

A variety of other neurally guided program induction algorithms have been proposed \cite{heapsearch, tree_diffusion, robustfill}. To the best of our knowledge, GridCoder differs from these algorithms mainly in the neural architecture used and in the fact that we lean heavily into the auto-regressive sequence output as the basis for enumerating the probability space. 

Most approaches use a neural network that outputs probabilities for each of the n-grams that can be formed by their DSL, relying instead more heavily on search than learning. Learning bigram probabilities, for example, means that the model learns to predict a probability for the children of a primitive $p_i$ conditionally on the choice of $p_i$ (and $p_i$ alone). Our auto-regressive  approach, instead, learns the probability over a child node given the entire program sequence (and the encoder embedding) so far.

As an illustration of this significant distinction, \citet{alford2021neurosymbolic} experimented on using DreamCoder on ARC-AGI, and concluded that the main bottleneck was the search efficiency. In our experiments with GridCoder, the main bottleneck was not the search efficiency (since the neural network was very good at identifying program structure), but its challenges at generalizing to unseen program structures (see the Discussion section for more details).

\section{Conclusion}

In summary, we compared three different program induction paradigms: LGS, LPS (specifically GridCoder) and LTS. Given the experimentation and analysis on GridCoder, we are confident that scaling up the DSL and training data could make it reach a performance that is competitive with the current top solutions (unless we later discover that these approaches explicitly solve the structural generalization problem instead of training on a dataset that unknowingly includes solutions to the hidden test set).

However, given its fundamental challenges with structural generalization, we are not confident that the LPS approach could solve ARC-AGI in any true sense. Instead, preliminary experimentation on the LTS approach suggests that GridCoder could be extended via execution-guided search to solve these generalization obstacles. This will be the focus of our future work.

\bibliography{aaai25}

\appendix

\section{Appendix - Description of the DSL}

The DSL revolves around the concept of a Grid, which is a custom class that contains the following information about a grid:
\begin{itemize}
    \item Its pixels, represented as a list of (X, Y, color) tuples.
    \item A utility method that converts from the pixels list to a 2D grid representation (a tuple of tuples where each value is the color at that position).
    \item Its width.
    \item Its height.
    \item Its upper-left corner position in the outer Grid. Default: (0, 0).
\end{itemize}

\paragraph{DSL Version 1:}

\begin{enumerate}
    \item \texttt{set\_fg\_color1} to \texttt{set\_fg\_color9}: set all the non-zero pixels to color 1 to 9.
    \item \texttt{shift\_left}: translate pixels once to the left.
    \item \texttt{shift\_right}: translate pixels once to the right.
    \item \texttt{shift\_up}: translate pixels once upward.
    \item \texttt{shift\_down}: translate pixels once downward.
    \item \texttt{vmirror}: flip the grid vertically.
    \item \texttt{hmirror}: flip the grid horizontally.
    \item \texttt{rot90}: rotate the grid 90 degrees.
    \item \texttt{rot180}: rotate the grid 180 degrees.
    \item \texttt{rot270}: rotate the grid 270 degrees.
    \item \texttt{tophalf}: keep only the top half of the grid.
    \item \texttt{bottomhalf}: keep only the bottom half of the grid.
    \item \texttt{lefthalf}: keep only the left half of the grid.
    \item \texttt{righthalf}: keep only the right half of the grid.
    \item \texttt{symmetrize\_left\_around\_vertical}: copy the left half to the right half, in a mirroring fashion.
    \item \texttt{symmetrize\_right\_around\_vertical}: copy the right half to the left half, in a mirroring fashion.
    \item \texttt{symmetrize\_top\_around\_horizontal}: copy the top half to the bottom half, in a mirroring fashion.
    \item \texttt{symmetrize\_bottom\_around\_horizontal}: copy the bottom half to the top half, in a mirroring fashion.
    \item \texttt{upscale\_horizontal\_by\_two}: duplicate pixels once, horizontally.
    \item \texttt{upscale\_vertical\_by\_two}: duplicate pixels once, vertically.
    \item \texttt{upscale\_by\_two}: duplicate pixels once, vertically and horizontally.
    \item \texttt{gravitate\_right}: stack the pixels to the right, as if they had fallen rightward.
    \item \texttt{gravitate\_left}: stack the pixels to the left, as if they had fallen leftward.
    \item \texttt{gravitate\_up}: stack the pixels at the top, as if they had fallen upward.
    \item \texttt{gravitate\_down}: stack the pixels at the bottom, as if they had fallen downward.
    \item \texttt{gravitate\_left\_right}: stack the leftmost pixels to the left, and the rightmost pixels to the right.
    \item \texttt{gravitate\_top\_down}: stack the top-half pixels at the top, and the bottom-half pixels at the bottom.
    \item \texttt{topthird}: keep only the top third of the grid.
    \item \texttt{vcenterthird}: keep only the middle third of the grid (vertically).
    \item \texttt{bottomthird}: keep only the bottom third of the grid.
    \item \texttt{leftthird}:  keep only the leftmost third of the grid.
    \item \texttt{hcenterthird}:  keep only the middle third of the grid (horizontally).
    \item \texttt{rightthird}: keep only the rightmost third of the grid.
    \item \texttt{cellwiseOR}: 2-argument primitive that takes in two grids of the same shape, and merges them into one by applying OR logic to the foreground pixels (prioritizing the first argument's pixel color if both grids have a foreground pixel at a certain position).
    \item \texttt{cellwiseAND}: 2-argument primitive that takes in two grids of the same shape, and merges them into one by applying AND logic to the foreground pixels (using the first argument's pixel color).
    \item \texttt{cellwiseXOR}: 2-argument primitive that takes in two grids of the same shape, and merges them into one by applying XOR logic to the foreground pixels.
    \item \texttt{cellwiseDifference}: 2-argument primitive that takes in two grids of the same shape, and merges them into one by setting the target grid cells the same color as the first argument's foreground pixels when the second argument does not contain a foreground pixel at that position.
    \item \texttt{cellwiseNOR}: 2-argument primitive that takes in two grids of the same shape, and places a foreground pixel of the same color as the dominant color in the first argument if neither grid has a foreground pixel at a given position.
    \item \texttt{vconcat}: 2-argument primitive that vertically concatenates two grids.
    \item \texttt{hconcat}: 2-argument primitive that horizontally concatenates two grids.
    \item \texttt{color\_change}: 3-argument primitive that takes as input a grid, and two integers representing colors. Every pixel in the grid that is of the color of the first integer gets converted to the color of the second integer. At search time, the integer values get resolved by an enumerative search built into the program evaluation logic.
    \item \texttt{invert\_colors}: Finds the least dominant color and the most dominant color, and swaps them in the grid.
    \item \texttt{first\_quadrant}: Return the top-left quadrant of the grid.
    \item \texttt{second\_quadrant}: Return the top-right quadrant of the grid.
    \item \texttt{third\_quadrant}: Return the bottom-left quadrant of the grid.
    \item \texttt{fourth\_quadrant}: Return the bottom-right quadrant of the grid.
    \item \texttt{hfirstfourth}: Returns the leftmost fourth of the grid (horizontal splitting).
    \item \texttt{hsecondfourth}: Returns the middle-left fourth of the grid (horizontal splitting).
    \item \texttt{hthirdfourth}: Returns the middle-right fourth of the grid (horizontal splitting).
    \item \texttt{hlastfourth}: Returns the rightmost fourth of the grid (horizontal splitting).
    \item \texttt{vfirstfourth}: Returns the topmost fourth of the grid (vertical splitting).
    \item \texttt{vsecondfourth}: Returns the middle-top fourth of the grid (vertical splitting).
    \item \texttt{vthirdfourth}: Returns the middle-bottom fourth of the grid (vertical splitting).
    \item \texttt{vlastfourth}: Returns the bottommost fourth of the grid (vertical splitting).
    \item \texttt{duplicate\_top\_row}: duplicates the rop row.
    \item \texttt{duplicate\_bottom\_row}: duplicates the bottom row.
    \item \texttt{duplicate\_left\_column}: duplicates the left column.
    \item \texttt{duplicate\_right\_column}: duplicates the right column.
    \item \texttt{remove\_outline}: removes the top row, bottom row, left column, right column.
    \item \texttt{shear\_grid\_left}: from the bottom up of the grid, it gradually shifts left the pixels by an increment that increases by 1 each row.
    \item \texttt{shear\_grid\_right}: from the bottom up of the grid, it gradually shifts right the pixels by an increment that increases by 1 each row.
    \item \texttt{shear\_grid\_zigzag}: from the bottom up of the grid, it  shifts the pixels horizontally by an increment that cycles between [-1, 0, +1].
    \item \texttt{insert\_outline}: it adds a black row at the top and bottom, and a black column at the left and the right.
    \item \texttt{get\_major\_pixel}: finds the most frequent color in the grid, and outputs a 1x1 grid of that color.
    \item \texttt{get\_minor\_pixel}: finds the least frequent color in the grid, and outputs a 1x1 grid of that color.
    \item \texttt{upscale\_by\_three}: upsamples the grid by a factor of three.
\end{enumerate}

\paragraph{DSL Version 2 - added primitives:}

\begin{enumerate}
    \item \texttt{get\_objects1}: returns a list of Grids, each representing an object. This variant is good at isolating rectangular shapes (filled or empty) from noisy backgrounds. 
    \item \texttt{get\_objects2}: returns a list of Grids, each representing an object. This variant is good at finding four-way (up, down, left, right) adjacent objects against a black background.
    \item \texttt{get\_objects3}:returns a list of Grids, each representing an object. This variant is good at finding eight-way adjacent objects against a black background.
    \item \texttt{get\_objects4}:returns a list of Grids, each representing an object. This variant is good at finding four-way (up, down, left, right) adjacent objects against of any color, as long as the object's color is uniform.
    \item \texttt{get\_objects5}:returns a list of Grids, each representing an object. This variant is good at finding eight-way adjacent objects against of any color, as long as the object's color is uniform.
    \item \texttt{get\_objects6}:returns a list of Grids, each representing an object. This variant is good at isolating shapes that are potentially disjoint, but clustered in a grid-like pattern.
    \item \texttt{compress\_objects\_linear}: this generates an output grid from a list of objects by concatenating them to each other in horizontal or vertically fashion, automatically determined from their relative positions.
    \item \texttt{compress\_objects\_quad}: this generates an output grid from a list of objects by concatenating them to each other in a rectangular fashion (mostly 2x2, 2x3, 3x2 and 3x3 patterns), automatically determined from their relative positions.
    \item \texttt{compress\_objects\_quad\_pad}: similar to \texttt{compress\_objects\_quad}, but introduces one row and column of black padding between the concatenated objects.
    \item \texttt{for\_each}: 2-argument primitive. The first argument is a list of objects to which to apply a function. The second argument is the lambda function to apply to each object. The input argument of that lambda function must be a Grid type.
    \item \texttt{apply\_to\_grid}: 2-argument primitive. The first argument is the original grid itself, and the second argument is the list of transformed objects to overlay onto the grid (blanking out the previous objects).
    \item \texttt{cellwise\_OR\_list}: like \texttt{cellwise\_OR}, but takes a list of two or more objects, instead of strictly only taking two arguments.
    \item \texttt{get\_pixels}: returns a list of integers representing all the colors of the pixels in the grid (this is not a set, but a list, so it can be used to count the number of pixels in an object, for example).
    \item \texttt{get\_object\_size}: returns an integer, the number of pixels in a Grid instance.
    \item \texttt{count}: counts the number of elements in a list.
\end{enumerate}

\paragraph{DSL Version 3 - added primitives:}

\begin{enumerate}
    \item \texttt{filter\_largest}: 2-argument primitive that takes as input a list of objects and a list of integer values. It returns a list of objects in which the object that has the largest associated integer value is removed.
    \item \texttt{filter\_smallest}: 2-argument primitive that takes as input a list of objects and a list of integer values. It returns a list of objects in which the object that has the smallest associated integer value is removed.
    \item \texttt{keep\_largest}: 2-argument primitive that takes as input a list of objects and a list of integer values. It returns the object from the list that has the largest associated integer value.
    \item \texttt{keep\_smallest}: 2-argument primitive that takes as input a list of objects and a list of integer values. It returns the object from the list that has the smallest associated integer value. 
    \item \texttt{is\_h\_symmetrical}: returns True is the grid is horizontally symmetrical, False otherwise.
    \item \texttt{is\_v\_symmetrical}: returns True is the grid is vertically symmetrical, False otherwise.
    \item \texttt{logical\_not}: takes as input a Boolean value, and returns its negation.
    \item \texttt{keep\_boolean}: 2-argument primitive that takes as input a list of objects and a list of Boolean values. It returns the first object from the list that has True as its associated value.
    \item \texttt{filter\_boolean}: 2-argument primitive that takes as input a list of objects and a list of Boolean values. It returns a list of objects in which the objects whose associated value is True are removed.
\end{enumerate}

\end{document}